\pdfoutput=1
\documentclass[11pt]{article}

\usepackage[final]{acl}
\usepackage{times}
\usepackage{latexsym}
\usepackage[T1]{fontenc}    
\usepackage[utf8]{inputenc}
\usepackage{microtype}
\usepackage{inconsolata}
\usepackage{graphicx}
\usepackage{booktabs}
\usepackage{tabularx}
\usepackage{url}
\usepackage{array}
\usepackage{multirow}
\usepackage{ragged2e}
\usepackage{caption}

\title{Benchmark Profiling: Mechanistic Diagnosis of LLM Benchmarks}

\author{
  \textbf{Dongjun Kim$^{1}$, Gyuho Shim$^{1}$, Yongchan Chun$^{1}$, Minhyuk Kim$^{1}$,} \\
  \textbf{Chanjun Park$^{2\dagger}$, Heuiseok Lim$^{1\dagger}$} \\
  $^{1}$Department of Computer Science and Engineering, Korea University \\
  $^{2}$School of Software, Soongsil University \\
  \texttt{\{junkim100, gjshim, cyc9805, mhkim0929, limhseok\}@korea.ac.kr} \\
  \texttt{chanjun.park@ssu.ac.kr}
}

\begin{document}
\maketitle

\begingroup
  \renewcommand{\thefootnote}{\fnsymbol{footnote}}
  \footnotetext[2]{Corresponding authors.}
\endgroup

\begin{abstract}
Large Language Models are commonly judged by their scores on standard benchmarks, yet such scores often overstate real capability since they mask the mix of skills a task actually demands. For example, ARC is assumed to test reasoning, while HellaSwag is designed to evaluate commonsense. However, we lack a systematic way to verify if these benchmarks actually measure these labels. We introduce \textsc{Benchmark Profiling}, a diagnostic framework that decomposes benchmark performance into ten cognitively grounded abilities. The method combines gradient-based importance scoring with targeted parameter ablation to compute an \textit{Ability Impact Score} (AIS) that quantifies how much each ability contributes to a model's success on a given benchmark. Profiling three instruction-tuned models across ten widely used benchmarks yields four key findings: (i) most benchmarks draw on several abilities rather than one, (ii) datasets with similar labels rely on distinct ability mixtures, (iii) code-generation benchmarks reward broad, multi-skill improvement and thus show only modest gains from narrow domain-specific fine-tuning, and (iv) abilities irrelevant to the task could negatively affect performance. \textsc{Benchmark Profiling} therefore explains why performance gains do not always translate into user-perceived competence and offers a transparent tool for benchmark audit and model interpretability. The code is available on \url{https://github.com/junkim100/Benchmark-Profiling}
\end{abstract}

\section{Introduction}

Modern evaluations of Large Language Models (LLMs) depend heavily on standardized benchmarks designed to test capabilities like reasoning, commonsense, and knowledge \citep{liang2022holistic, cobbe2021training, zellers2019hellaswag}. While these benchmarks provide quantitative measures of performance, a growing body of evidence suggests a discrepancy between high scores on automated metrics and the qualities humans value in LLM interaction. For instance, models optimized for benchmarks can sometimes produce outputs that are misaligned with human preferences, as evidenced by the mismatched rankings between platforms like the Open LLM Leaderboard \citep{open-llm-leaderboard-v2} and the Chatbot Arena LLM Leaderboard \citep{chiang2024chatbot}. This misalignment raises a critical concern: \textbf{current benchmarks may not accurately measure the abilities they claim to assess}, undermining their reliability as indicators of true model competence.

\begin{figure}[t]
\centering
\includegraphics[width=\columnwidth]{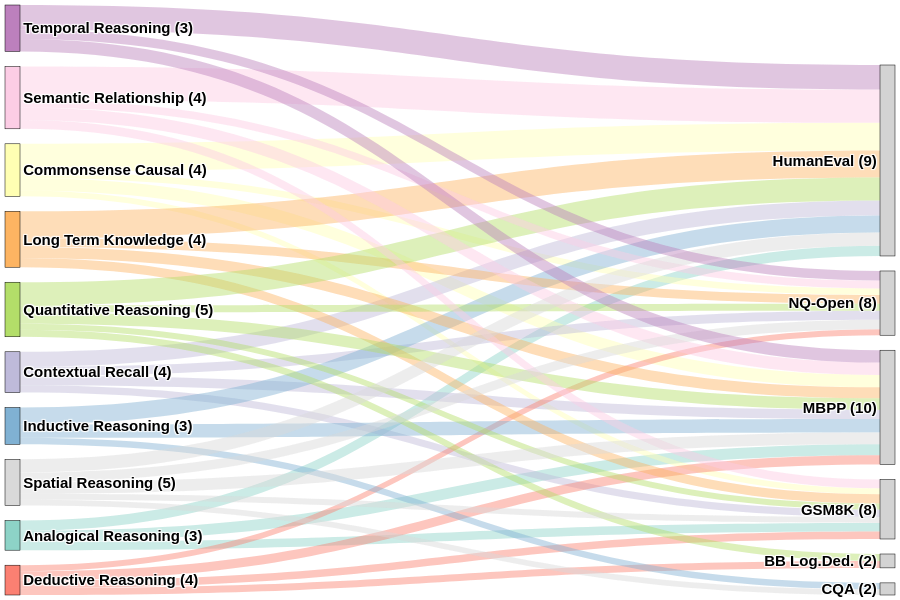}
\caption{
Top ability-benchmark links for Llama-3.1-8B-Instruct derived from its Benchmark Profile (ribbons shown only where $\mathrm{AIS}>0.05$; ribbon width $\propto$ impact).
}
\label{fig:chord}
\end{figure}

The core issue lies in the ambiguity of benchmark design. High accuracy scores on benchmarks are often taken as direct proof that a model possesses the high-level ability suggested by the benchmark's label (e.g., \textit{math} or \textit{commonsense}), despite a lack of rigorous verification \citep{eriksson2025can}. In reality, models might exploit dataset artifacts or memorize patterns to achieve high scores without genuine understanding \citep{mccoy2019right, geva2021aristotle}. Without knowing what benchmarks truly measure, we cannot reliably improve models or design evaluations that reflect real-world requirements \citep{bowman2021measuring}.

To address this, we introduce \textsc{Benchmark Profiling}, a methodology that systematically diagnoses the functional abilities required by LLM benchmarks. By defining 10 operationalized abilities (e.g., Deductive Reasoning, Contextual Recall) derived from established models of human intelligence \citep{carroll1993human}, we create measurement criteria that reflect both computational performance and the cognitive dimensions humans intuitively value in real-world interactions. This approach directly tackles the \textit{Performance-Perception Paradox}, where models dominate benchmarks yet underwhelm users, by ensuring evaluations test the same competencies people assess when judging capability \citep{kyllonen2021taxonomy}. Bridging this gap, our profiles reveal whether "high-scoring" models truly exhibit the abilities users expect from labels like \textit{math} or \textit{commonsense}. The \textsc{Benchmark Profiling} framework measures how much each ability actually contributes to a model's success on each benchmark, using targeted parameter ablation and our proposed Ability Impact Score (AIS). This approach produces diagnostic profiles that reveal the true combination of abilities required for high performance on every benchmark.

\begin{figure*}[t]
\centering
\includegraphics[width=\linewidth]{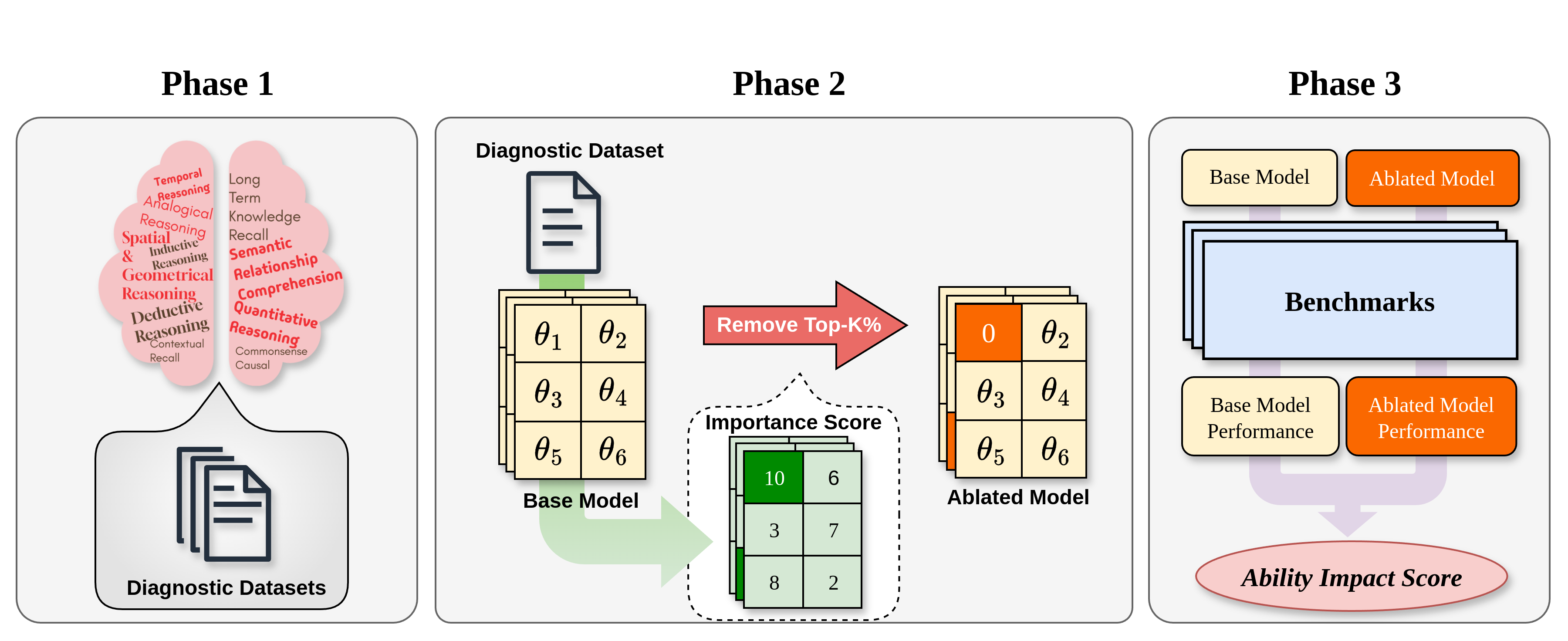}
\caption{
Three-step pipeline of \textsc{Benchmark Profiling}.
\textbf{Left:} We define ten cognitively motivated abilities and create a dedicated diagnostic dataset for each one.
\textbf{Middle:} Using the diagnostic dataset, we rank the base model's parameters by gradient-based importance, and zero out (orange) the top~$k$ percent associated with that ability.
\textbf{Right:} We run the original and ability-ablated models on downstream benchmarks. Their task accuracies yield the \emph{Ability Impact Score (AIS)}, which quantifies how strongly the benchmark depends on the ablated ability.}
\label{fig:method_overview}
\end{figure*}

\section{Related Work}
\label{sec:related_work}
\subsection*{The Benchmarking Paradigm in LLM Evaluation}

Large Language Models (LLMs) are predominantly evaluated through standardized benchmarks like MMLU \citep{hendrycks2021measuring}, HellaSwag \citep{zellers2019hellaswag}, and TruthfulQA \citep{lin2021truthfulqa}, which quantify performance on tasks such as commonsense reasoning, mathematical problem-solving, and factual accuracy. While these benchmarks have driven rapid progress via leaderboards, their limitations in capturing real-world competence and human-aligned abilities have become increasingly apparent. Critiques highlight issues such as dataset contamination, prompt sensitivity, and the prevalence of shortcut learning, where models exploit artifacts rather than demonstrating genuine understanding \citep{mccoy2019right, geva2021aristotle, bowman2021measuring}.
% \subsection*{Limitations of Current Evaluation Paradigms}

A growing body of research reveals systemic flaws in the benchmarking paradigm. Studies demonstrate that models often achieve high scores through memorization or spurious correlations, failing to exhibit robust reasoning or adaptability \citep{banerjee2024vulnerabilitylanguagemodelbenchmarks, oh-etal-2024-generative}. For instance, inherent limitations, such as overfitting to narrow metrics, and a lack of theoretical justification for real-world relevance, are shown in benchmarks like MMLU containing mislabeled or irrelevant questions \citep{fodor2025linegoesupinherent}. These critiques align with observations of the \textit{performance-perception paradox}, a term we introduce to describe the disconnect between benchmark-driven progress and the abilities users intuitively expect from LLMs in practical applications. Further analyses reveal that benchmarks often inadequately measure reasoning depth, exhibit cultural biases, and struggle with adversarial robustness \citep{McIntosh_2025}, underscoring the need for evaluations grounded in human-aligned competencies.

\subsection*{Mechanistic Interpretability in Language Models}

Mechanistic interpretability seeks to reverse engineer neural networks by mapping their internal computations to human-understandable algorithms and concepts, aiming for a granular, causal understanding of model behavior \citep{bereska2024mechanistic}. This approach distinguishes itself through its ambition to \textit{completely specify a neural network's computation}, enabling researchers to diagnose how models encode knowledge and execute tasks. In language models, mechanistic studies have uncovered computational mechanisms for syntactic processing \citep{hewitt-manning-2019-structural} and factual recall \citep{elhage2022superposition}, often through techniques like probing and ablation. Central to this effort are gradient-based importance scores, which quantify parameter contributions to task performance by analyzing the sensitivity of the loss function to perturbations \citep{molchanov2019importance, Michel2019sixteen}. These methods provide a practical means to identify critical parameters for specific abilities, bridging the gap between reverse engineering and actionable diagnostics.

Recent advances in mechanistic interpretability provide a foundation for critically assessing benchmark validity. While earlier studies focused on model behaviors \citep{yu2024interpretingarithmeticmechanismlarge,zhang2024unveilinglinguisticregionslarge, stolfo2023mechanisticinterpretationarithmeticreasoning} (e.g., shortcut learning in arithmetic tasks \citep{geva2021aristotle}), our work introduces a novel application of these insights to evaluate whether benchmarks genuinely measure the abilities they claim to assess. By operationalizing cognitive dimensions and quantifying their impact through targeted parameter ablation, we demonstrate how mechanistic tools can diagnose mismatches between benchmark requirements and human-aligned competencies. This approach addresses a key gap in prior critiques, which identified limitations but lacked methodologies to systematically evaluate benchmark validity \citep{fodor2025linegoesupinherent, McIntosh_2025}, positioning mechanistic interpretability as a critical tool for advancing evaluation frameworks that reflect real-world capabilities.

\begin{table*}[t]
\centering
\small
\renewcommand{\arraystretch}{1.2}
\begin{tabularx}{\linewidth}{l l X}
\toprule[1.5pt]
\textbf{Abbr.} & \textbf{Ability} & \textbf{Operationalization in Diagnostic Dataset}\\
\midrule
\textit{Ana}   & \textbf{Analogical Reasoning} &
Present an analogy or proportional pair (e.g.\ \textit{A:B :: C:?}) and ask which option best completes the relationship.  
Distractors ensure success requires mapping the underlying relation rather than surface word similarity.\\[0.6ex]

\textit{Com}  & \textbf{Commonsense \& Causal Reasoning} &
Give a short everyday vignette and ask for the most plausible cause, effect, or next event; items hinge on everyday causal plausibility, not memorized facts.\\[0.6ex]

\textit{Cxt}   & \textbf{Contextual Recall} &
Provide a brief passage, then ask for verbatim details or their conjunction without new inference, isolating short-term textual memory.\\[0.6ex]

\textit{Ded}   & \textbf{Deductive Reasoning} &
Present premises that logically entail one conclusion; decoy options violate at least one logical step, forcing rule-based inference.\\[0.6ex]

\textit{Ind}   & \textbf{Inductive Reasoning} &
Show a short pattern or sequence and ask the model to infer the governing rule and extrapolate, so only rule discovery generalizes.\\[0.6ex]

\textit{LTK}   & \textbf{Long-Term Knowledge Recall} &
Ask about stored factual knowledge (history, science, geography) absent from the prompt; items use low-frequency facts to reduce chance memorization from local context.\\[0.6ex]

\textit{Quant} & \textbf{Quantitative Reasoning} &
Pose a word problem with numerical data requiring arithmetic or counting; multi-step reasoning and distractor numbers discourage pattern matching.\\[0.6ex]

\textit{Sem}   & \textbf{Semantic Relationship Comprehension} &
Give a passage with several entities and ask about their roles or relations (e.g.\ part-whole, managerial hierarchy); questions test explicit and implicit links, not mere co-occurrence.\\[0.6ex]

\textit{Spat}  & \textbf{Spatial \& Geometrical Reasoning} &
Describe spatial layouts or geometric facts, then ask about positions, directions, shapes, or distances; requires constructing a mental map or performing shape-based deductions.\\[0.6ex]

\textit{Temp}  & \textbf{Temporal Reasoning} &
Present events with time markers (dates, times, order words) and ask about sequence, simultaneity, or duration; items mix explicit and implicit cues to test chronology.\\
\bottomrule[1.5pt]
\end{tabularx}
\caption{Operationalized abilities and their abbreviations used in the \textsc{Benchmark Profiling} framework.}
\label{tab:cognitive_abilities}
\end{table*}

\section{Methodology}
\label{sec:cbp_methodology}

\textsc{Benchmark Profiling} is a systematic methodology designed to diagnose the ability composition of LLM evaluation benchmarks. It quantifies the dependence of benchmarks on a predefined set of fundamental operationalized abilities by measuring the impact of selectively ablating ability-specific parameters within an LLM. The methodology comprises three main phases:

\subsection*{Phase 1: Defining Abilities}

A cornerstone of \textsc{Benchmark Profiling} is establishing a set of well-defined, fundamental abilities that serve as the diagnostic criteria. To address the \textit{Performance-Perception Paradox} where models excel on benchmarks yet underperform in human-aligned contexts, we ground these criteria in established cognitive science frameworks \citep{beinborn2024cognitive}. By building on taxonomies like Cattell-Horn-Carroll (CHC) theory \citep{carroll1993human}, which describes human cognitive abilities such as fluid reasoning and working memory, we ensure our operationalized abilities reflect dimensions humans intuitively recognize as markers of intelligence. This human-centric foundation bridges the gap between benchmark scores and the competencies users expect LLMs to exhibit in real-world interactions. In designing these 10 abilities found in Table~\ref{tab:cognitive_abilities}, we balance theoretical robustness with practical relevance by adapting cognitive science principles to the context of LLM evaluation tasks, ensuring that each ability is both grounded in human cognition and directly applicable to benchmarking modern language models. While inspired by human cognition, these terms refer to specific, operationalized functional capacities within the LLM architecture. Detailed definitions are in Appendix~\ref{app:ability_definitions}.

For each defined ability \(a\), a diagnostic dataset \(D_a\) is created (2000 MCQs per ability in this work) which is designed to specifically measure that ability. Crucially, these datasets are validated in Section~\ref{sec:dataset_validation}, and creation details are in Appendix~\ref{app:dataset_creation}.

\subsection*{Phase 2: Identifying Abilities}
This phase identifies specific components within the LLM, that are responsible for each defined ability. Within a chosen LLM (\(\Theta\)), parameters critical for each dataset \(D_a\) are identified.

\paragraph{Importance Scoring} We compute gradient-based importance scores \(I^a_j(\theta)\) for each parameter \(\theta_j\) using a first-order Taylor approximation of the loss \(L(D_a, \theta)\) on dataset \(D_a\) \citep{molchanov2019importance, Michel2019sixteen}.
    
\begin{equation}
\label{eq:importance}
I^a_j(\theta) \approx \left| \frac{\partial L(D_a, \theta)}{\partial \theta_j} \cdot \theta_j \right|
\end{equation}

Gradients are obtained via fine-tuning on \(D_a\). This fine-tuning is performed solely to facilitate accurate gradient estimation. The resulting model state is \textbf{discarded}.

\paragraph{Parameter Selection} MLP layer parameters are ranked by \(I^a_j(\theta)\), and the top-k\% are selected as the parameter subset associated with ability \(a\). For each ability \(a\), an ablated model \(\Theta^{a}\) is created by taking the original model \(\Theta\) and setting the value of identified top-k\% MLP parameters for ability \(a\) to zero. Preliminary experiments revealed that restricting ablations to MLP weights yields the clearest ability-specific signal with minimal collateral damage; see Section~\ref{sec:mlp_validation} for details.

\subsection*{Phase 3: Benchmark Profiling}
This phase involves evaluating baseline and ablated models on target benchmarks, calculating the Ability Impact Score (AIS) to normalize performance changes, and constructing the Benchmark Profile from these AIS values.

The original model \(\Theta\) and each ablated model \(\Theta^{a}\) are evaluated on target benchmarks \(b\). Let baseline performance be \(P_b(\Theta)\) and ablated performance be \(P_b(\Theta^{a})\).

To quantify benchmark reliance on each ability, we define the Ability Impact Score (AIS) for ability \(a\) on benchmark \(b\), measuring the proportion of performance loss relative to the model's baseline improvement over chance:

\begin{equation}
\label{eq:ais}
\mathrm{AIS}^{a}_b = \frac{P_b(\Theta) - P_b(\Theta^{a})}{P_b(\Theta) - P^\mathrm{chance}_b}
\end{equation}

where \(P^\mathrm{chance}_b\) is chance-level performance for benchmark \(b\). An AIS near 1 indicates strong dependence, while an AIS near 0 suggests little or no reliance. A negative AIS means that performance actually improves after the ability is ablated, signaling that the ability can be detrimental for that benchmark.

The calculated AIS values (\(\mathrm{AIS}^{a}_b\)) are organized into the \textbf{Benchmark Profile}, providing a quantitative summary of each benchmark's measured reliance on the defined operationalized abilities.

\begin{figure*}[t]
\centering
\includegraphics[width=\linewidth]{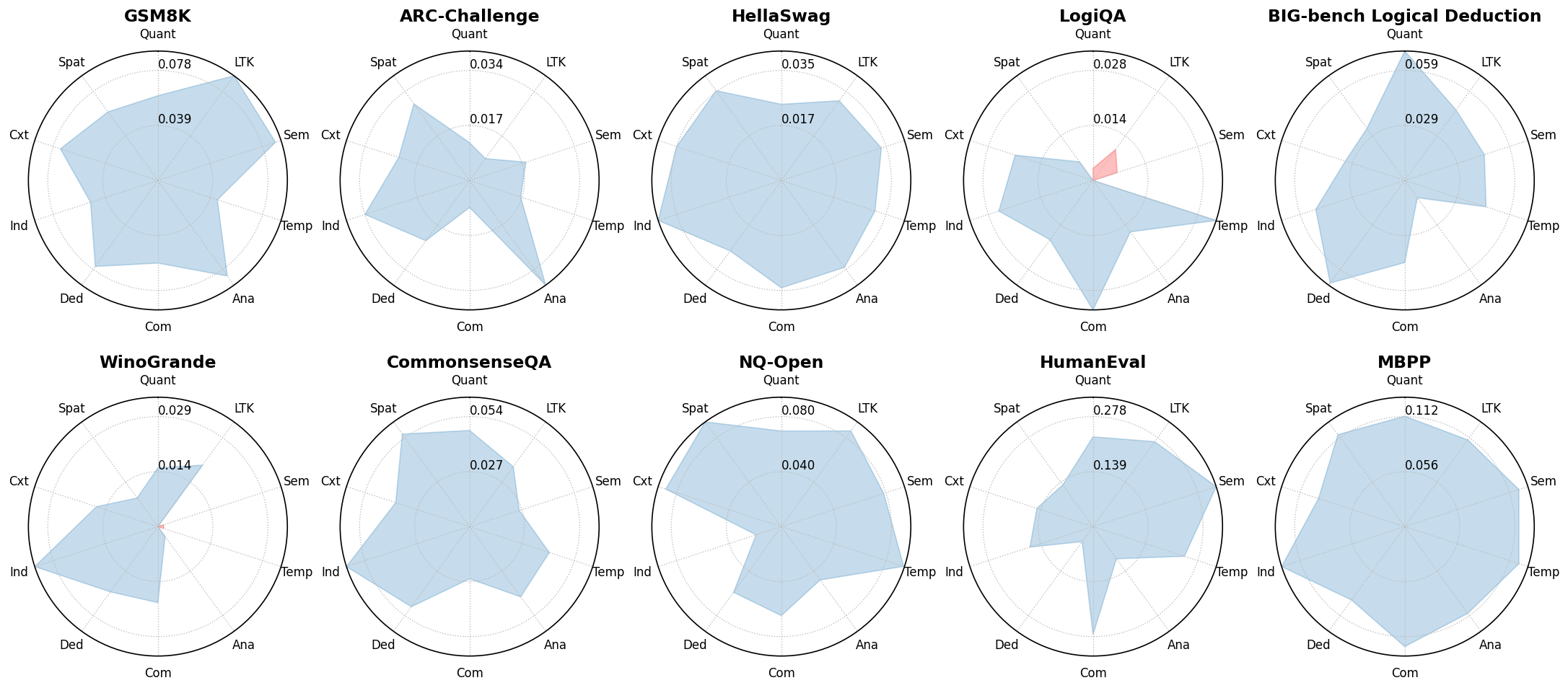} % ten-panel radar montage
\caption{
Ability Impact Score radar plots for the ten benchmarks profiled on Llama-3.1-8B-Instruct.  
Axes are labeled with the following abbreviated abilities. Blue and red shading indicates positive and negative AIS values.}
\label{fig:radar_llama}
\end{figure*}

\section{Experiments}
\label{sec:experiments}

This section details how \textsc{Benchmark Profiling} is applied to derive Ability Impact Scores (AIS) for a suite of standard benchmarks.  Section~\ref{sec:experimental_design} explains the procedure for pinpointing ability-specific parameters and computing AIS values.  Section~\ref{sec:experimental_setup} describes the experimental setup, including the language models, diagnostic datasets, and target benchmarks.  The Benchmark Profiles produced by these experiments are interpreted in Section~\ref{sec:main_results}.

\subsection{Experiment Design}
\label{sec:experimental_design}

Our goal is to quantify how strongly each benchmark in the curated suite depends on each of the ten operationalized abilities. For every ability we first rank model weights by gradient-based importance, then ablate the \emph{top} \(1.024\,\%\) of MLP parameters associated with that ability. A preliminary sweep across smaller and larger \(k\) values ranging from \(0.001\,\%\) to \(4.096\,\%\) showed that \(1.024\,\%\) is the smallest budget that produces a clear, ability-specific signal without inflicting unnecessary collateral damage on unrelated capabilities. Applying this threshold yields ten ability-ablated models, each of which selectively disrupts one functional component while leaving the rest of the network, and its fluency, largely intact.

The core aim of our experiment design is to systematically measure how much each benchmark in our curated suite depends on each of the 10 operationalized abilities. For each ability, we identify the most critical model parameters using gradient-based importance scores, then create an ablated model by zeroing out the top 1.024\% of MLP parameters associated with that ability. This process yields 10 ability-ablated models, each designed to selectively disrupt one functional component while leaving the rest of the model intact.

We systematically evaluated the baseline and ability-ablated models on our curated suite of 10 benchmarks, applying each benchmark's standard evaluation metric. For every ability-benchmark pair, we computed the AIS as the normalized performance drop relative to the model's improvement over chance, as formalized in Equation~\ref{eq:ais}. This yields the Benchmark Profile, which quantifies the functional dependence of each benchmark on each operationalized ability. The Benchmark Profile serves as the foundation for all subsequent analyses, providing a quantitative map of dependencies that we interpret in the following sections.

\subsection{Experimental Setup}
\label{sec:experimental_setup}

This section outlines the core components used in our experiments: the LLMs subjected to profiling, the diagnostic datasets developed to isolate specific abilities, and the suite of benchmarks selected for analysis, including details on their evaluation.

\paragraph{Models}
Our primary experiments leverage Llama-3.1-8B-Instruct \citep{vavekanand2024llama}, a widely recognized instruction-tuned model. To assess the generalizability of our findings, robustness checks replicate key analyses on two additional models: Qwen/Qwen2.5-7B-Instruct \citep{qwen2_2024} and mistralai/Mistral-7B-Instruct-v0.3 \citep{jiang2024mistral}. All models are used in their base precision (BF16) and evaluated using greedy decoding, consistent with common practices, unless otherwise specified by a benchmark's standard protocol.

\paragraph{Diagnostic Datasets}
We employ the 10 diagnostic datasets designed to target the operationalized abilities defined in Section~\ref{sec:cbp_methodology} and listed in Table~\ref{tab:cognitive_abilities}. Each dataset consists of 2000 4-choice Multiple-Choice Questions (MCQs), totaling 20,000 examples. These datasets were synthetically generated using the \texttt{o4-mini-2025-04-16} API via carefully crafted few-shot prompting strategies tailored to each ability. Detailed descriptions of the generation prompts and examples for each ability dataset are provided in Appendix~\ref{app:dataset_creation}. The validation of these datasets is presented in Section~\ref{sec:dataset_validation}.

\paragraph{Benchmark Details}
For the application of \textsc{Benchmark Profiling}, we selected the curated suite of 10 standard LLM evaluation tasks. This suite was chosen to encompass a variety of task formats and evaluation paradigms common in LLM assessment.

The selected benchmarks include several multiple-choice question-answering tasks: ARC-Challenge \citep{clark2018think} which uses a 4-choice format; HellaSwag \citep{zellers2019hellaswag}, also 4-choice, requiring sentence completion; WinoGrande \citep{sakaguchi2021winogrande}, a 2-choice pronoun resolution task; CommonsenseQA \citep{talmor2019commonsenseqa}, a 5-choice QA task; LogiQA \citep{liu2020logiqa}, a 4-choice QA over logical passages; and BIG-Bench Logical Deduction \citep{srivastava2022beyond}, a 5-choice task.

The suite also incorporates generation tasks. GSM8K \citep{cobbe2021training} requires generating a chain-of-thought leading to a final numerical answer, which is then matched for evaluation. Natural Questions Open (NQ-Open) \citep{kwiatkowski2019natural} is an open-domain QA task where short generated answers are evaluated by exact match. For coding, HumanEval \citep{chen2021evaluating} and MBPP (Mostly Basic Python Problems) \citep{austin2021program} require the model to generate Python code, which is then evaluated for functional correctness using a pass@1 metric.

To ensure consistency and facilitate reproducible evaluations across this diverse suite, we utilized the EleutherAI Language Model Evaluation Harness \citep{eval-harness} for executing the benchmark tasks and collecting performance metrics. For each benchmark, we adhere to its standard evaluation protocol and primary metric. These performance scores are subsequently used to calculate the AIS as defined in Section~\ref{sec:cbp_methodology} Phase 3. The chance-level performance $P^\mathrm{chance}_b$ for each benchmark, critical for the AIS calculation, is determined by its specific format (e.g., 0.25 for 4-choice MCQs, 0.5 for 2-choice, 0.2 for 5-choice, and 0 for generation tasks). This diverse set of task formats and evaluation approaches allows us to investigate how ability dependencies manifest across different interaction and assessment modalities.

\section{Main Results}
\label{sec:main_results}

This section interprets the benchmark profile of Llama-3.1-8B-Instruct. We visualize AIS the patterns with radar plots and compare cross-model similarity with Jensen--Shannon statistics. We then highlight four empirical observations, showing  (i) that popular benchmarks exercise multiple abilities rather than a single labeled skill, (ii) that seemingly related datasets often reward very different mixtures of abilities, (iii) that code-generation tasks demand the broadest spectrum of abilities and therefore penalize narrow fine-tuning, and (iv) that certain abilities can act as distractors on tightly constrained reasoning tasks. The remainder of the section presents the visual evidence and discusses each \textit{Key Finding} in detail. Detailed AIS matrix can be found in Table~\ref{tab:ais_main_detailed}.

\paragraph{Key Finding 1: Benchmarks Combine Multiple Abilities}
Figure~\ref{fig:radar_llama} reveals that every benchmark draws on a rich mixture of skills. HellaSwag and MBPP show a broad footprint, while WinoGrande, the narrowest profile, still includes more than one competency. \textbf{GSM8K} peaks in \textit{Long-Term Knowledge Recall} and \textit{Semantic Relationship}, while \textit{Quantitative Reasoning} is only moderate.  \textbf{ARC-Challenge} centers on \textit{Analogical} and \textit{Inductive Reasoning} with minimal \textit{Long-Term Knowledge Recall}. \textbf{LogiQA}, marketed as a logical reasoning benchmark, in fact leans most on \textit{Temporal Reasoning} and \textit{Commonsense Causal Reasoning}, with \textit{Deductive Reasoning} contributing only modestly. These composite patterns confirm that task labels such as \textit{math} or \textit{logic} under-specify what is really being measured. 

\paragraph{Key Finding 2: Benchmarks with Similar Labels Test Different Abilities}
Figure \ref{fig:radar_llama} compares two question-answering datasets that are often grouped under \textit{knowledge QA} yet rely on markedly different skill mixes. \textbf{CommonsenseQA}, a 5-choice multiple-choice benchmark, peaks in \textit{Inductive Reasoning} and draws secondary support from \textit{Deductive Reasoning}, \textit{Spatial Reasoning}, \textit{Quantitative Reasoning}, and \textit{Analogical Reasoning}. In contrast, \textbf{Natural Questions Open} (NQ-Open), an open-ended retrieval task, scores highest on \textit{Temporal Reasoning}, \textit{Spatial Reasoning}, \textit{Semantic Relationship}, \textit{Long-Term Knowledge}, and \textit{Contextual Recall}.

These divergent ability footprints translate into sharply different accuracies: Llama-3.1-8B-Instruct answers 77.1\% of CommonsenseQA items correctly yet attains only a 17.9\% exact-match rate on NQ-Open; Qwen2.5-7B-Instruct shows a similar contrast (82.7\% vs. 4.7\%). Even after accounting for the easier multiple-choice format of CommonsenseQA, the gap remains large. Such crossed scores illustrate how a model can excel on one \textit{knowledge QA} benchmark while struggling on another that depends on a different blend of abilities, underscoring the diagnostic value of \textsc{Benchmark Profiling}.

\paragraph{Key Finding 3: Code Benchmarks Demand Broad Skill Sets}
The two bottom-right panels of Figure~\ref{fig:radar_llama} show that \textbf{HumanEval} and \textbf{MBPP} produce the largest AIS values, indicating that success depends on many abilities at once. HumanEval is driven most by \textit{Semantic Relationship}, aligning with the need to interpret function specifications precisely, whereas MBPP lights up almost every axis forming an almost complete disk.

The wide spread of AIS values for MBPP aligns with findings that coding datasets inherently correlate with multiple reasoning abilities due to their structured, logic-driven nature \citep{zhang2024unveilingimpactcodingdata}. This mutual reinforcement has been evident in training dynamics: models exposed to code data not only excel at programming tasks but also exhibit enhanced performance on mathematical and logical reasoning benchmarks \citep{ma2023trainingstagedoescode, tao2024crystalilluminatingllmabilities}. This correlation between code and the broad spectrum of reasoning abilities explains why MBPP’s profile lights up nearly every ability axis. The structured syntax and semantic precision required in coding tasks act as a basis for multitask learning, reinforcing skills like deductive reasoning and contextual recall that are critical for both programming and general problem-solving.

\begin{figure}[t]
\centering
\includegraphics[width=\linewidth]{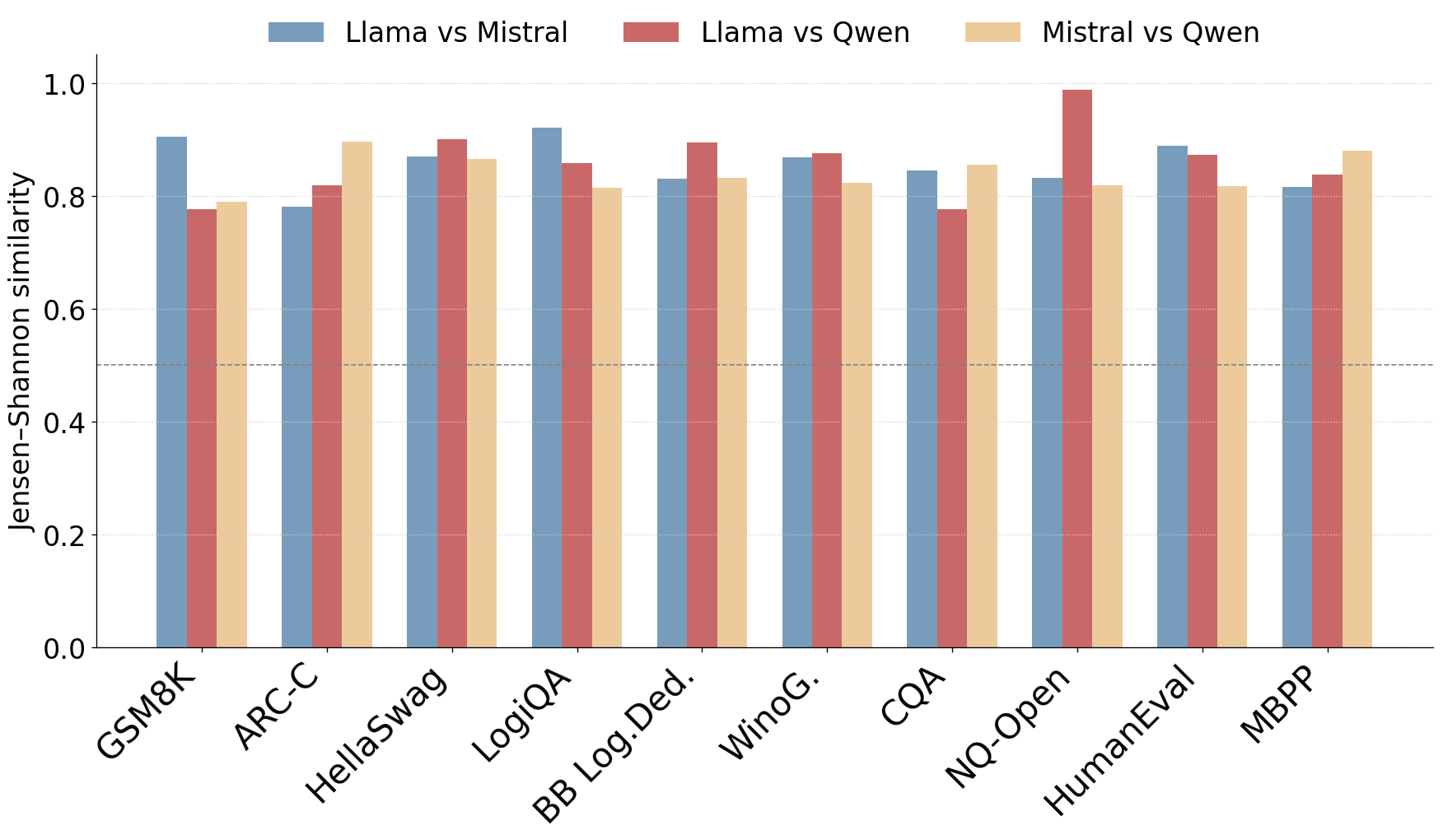} % Jensen--Shannon bar plot
\caption{Jensen--Shannon Similarity after min-max normalization.  Each bar compares two models on a single benchmark.}
\label{fig:js_similarity}
\end{figure}

\paragraph{Key Finding 4: Irrelevant Abilities Can Hurt Performance}
Figure~\ref{fig:radar_llama} exposes a small but consistent pocket of \emph{negative} AIS values shown in red: ablating \textit{Long-Term Knowledge}, \textit{Semantic Relationship}, or \textit{Quantitative Reasoning} increases LogiQA accuracy by 1--2 percentage points, and removing \textit{Temporal Reasoning} or \textit{Semantic Relationship} yields a similar boost on WinoGrande.  
This pattern is absent from the other eight benchmarks, indicating that negative transfer emerges only when the dataset contains spurious cues that conflict with its core reasoning chain.  
LogiQA is explicitly constructed so that the correct answer is derivable \emph{only} from the supplied premises; introducing external factual recall or numerical heuristics therefore lures a model toward plausible-but-invalid distractors \citep{liu2020logiqa}.  
WinoGrande was adversarially re-balanced to neutralize superficial lexical biases, forcing systems to rely on fine-grained syntactic cues; augmenting the model with world knowledge or event-ordering heuristics re-introduces precisely the shortcut signals the benchmark was designed to suppress \citep{sakaguchi2021winogrande,mccoy2019right}.  
More broadly, multi-task learning research shows that adding tasks or features weakly correlated with the gold decision boundary can hurt generalization, a phenomenon known as \emph{negative transfer} \citep{zhang2022survey}.  
Recent mechanistic and robustness studies echo this observation, demonstrating that \emph{adding} external knowledge or auxiliary data can introduce spurious correlations that \emph{degrade} downstream logical-reasoning accuracy \citep{schuff2021does,compton2023more}.
Because the remaining benchmarks either reward those auxiliary abilities or embed them in ways that align with the task objective, ablating them provides no systematic benefit, hence negative AIS values appear only for LogiQA and WinoGrande.

\paragraph{Robustness Across Models}
After min-max-normalizing each benchmark column of the AIS matrix, we measure agreement with \emph{Jensen--Shannon Similarity (JSS)}, which is derived from the Jensen--Shannon 
Divergence. For two discrete distributions $p$ and $q$, it is expressed as:
\[
\mathrm{JSS}(p,q)=
1-\frac{1}{2}(\,\mathrm{D_{KL}}(p\parallel m)+\,\mathrm{D_{KL}}(q\parallel m))
\]
\noindent where $m=\frac{1}{2}(p+q)$ and $\mathrm{D_{KL}}$ is Kullback--Leibler divergence. By construction, $\mathrm{JSS}(p,q)\in[0,1]$, with higher values indicating greater similarity.

Figure \ref{fig:js_similarity} plots the pairwise similarities for all ten benchmarks.  Every bar is above the gray 0.5 reference line (range 0.53--0.89, mean 0.64), indicating that Llama-3.1-8B, Mistral-7B, and Qwen-2.5 share broadly consistent ability footprints despite architectural differences.

\begin{table}[t]
\centering
\renewcommand{\arraystretch}{1.3}
\resizebox{\linewidth}{!}{%
\begin{tabular}{lcccccc}
\toprule[1.5pt]
& \multicolumn{3}{c}{\textbf{GSM8K}} 
& \multicolumn{3}{c}{\textbf{NQ-Open}} \\
\cmidrule(lr){2-4}\cmidrule(l){5-7}
\textbf{Ablated Ability} &
\textbf{Base} & \textbf{MLP} & \textbf{All} &
\textbf{Base} & \textbf{MLP} & \textbf{All} \\
\midrule[1pt]
\textbf{Contextual Recall} &
\multirow{2}{*}{0.773} & 0.7354 & 0.1024 &
\multirow{2}{*}{0.1789} & 0.1202 & 0.0374 \\[0.2ex]
\textbf{Quantitative Reasoning} &
& 0.7422 & 0.0902 &
& 0.1357 & 0.0163 \\
\bottomrule[1.5pt]
\end{tabular}}
\caption{Top-1.024\% ablation applied either to MLP weights only
(\textit{MLP}) or to all weights (\textit{All}). Results are reported as exact-match accuracy.}
\label{tab:mlp_vs_all_layer_comparison}
\end{table}

\section{Validation of Methodology Components}
\label{sec:validation}

We validate our method in two ways, first by having human experts confirm that each diagnostic dataset genuinely targets its stated ability, and second by demonstrating that ablating only MLP parameters weakens the intended skills while preserving overall model fluency better than ablating all layers.

\subsection{Expert Evaluation of the Diagnostic Datasets}
\label{sec:dataset_validation}

To confirm that each prompt truly targeted its intended ability, we asked ten independent domain specialists to review a stratified sample of items from every ability category (annotator demographics and instructions in Appendix~\ref{app:human_eval}).  
Each expert saw the \emph{context}, \emph{question}, and \emph{answer} for every item, then  
(i) selected which of the ten ability labels best described the required skill and  
(ii) judged whether the item fit that definition. Experts selected the correct label in \textbf{92.2\%} of cases (individual ability accuracies: 74\%, 88\%, 92\%, 92\%, 92\%, 94\%, 94\%, 98\%, 98\%, 100\%), confirming that the items faithfully captured their intended skills.

\subsection{MLP-Only Ablation}
\label{sec:mlp_validation}

Zeroing attention weights as well as MLP weights quickly dismantles the self-attention pathways that bind tokens into a coherent context that supports fluent text. Table \ref{tab:mlp_vs_all_layer_comparison} contrasts the two ablation regimes on Llama-3.1-8B-Instruct for the \emph{Contextual Recall} and \emph{Quantitative Reasoning} abilities. The \textit{MLP-only} variant yields only modest accuracy drops, whereas the \textit{all-layer} variant slashes performance on every setting in the table, confirming that attention layer damage wipes out far more capability than is needed for diagnostic purposes. Appendix \ref{app:qualitative_mlp} shows samples that match these numbers: the MLP-only model stays fluent, whereas the all-layer model lapses into repetitive, incoherent text.

\section{Conclusion}
\label{sec:conclusion}

Current benchmark tasks often obscure which skills a language model actually employs during evaluation, making it difficult to know when a reported gain reflects robust problem-solving ability or an exploitable shortcut. We introduce \textsc{Benchmark Profiling}, a systematic framework that decomposes benchmark performance into ten operationalized abilities grounded in cognitive science.  By combining gradient-based importance scoring, targeted parameter ablations, and the Ability Impact Score, our method delivers an interpretable ability fingerprint for every benchmark-model pair.

Experiments applying \textsc{Benchmark Profiling} to widely used models and benchmarks uncovered patterns indicating that most benchmarks tap several underlying abilities, tasks with the same label often depend on different ability blends, code-generation benchmarks reward broad multi-skill competence rather than narrow domain tuning, and adding abilities a task does not truly demand can even reduce performance. These insights clarify why leaderboard gains sometimes fail to translate into practical capability.

\textsc{Benchmark Profiling} thus provides researchers and practitioners with transparent diagnostics, enabling better-aligned model evaluations, targeted improvements in model design, and more accurate interpretations of benchmark results.

\section*{Acknowledgements}
This work was supported by the Commercialization Promotion Agency for R\&D Outcomes (COMPA) grant funded by the Korean government (Ministry of Science and ICT) (2710086166). This work was supported by the Institute for Information \& Communications Technology Promotion (IITP) grant funded by the Korean government (MSIT) (RS-2024-00398115, Research on the reliability and coherence of outcomes produced by Generative AI). This research was supported by the Basic Science Research Program through the National Research Foundation of Korea (NRF), funded by the Ministry of Education (NRF-2021R1A6A1A03045425).

\section*{Limitations}
\label{sec:limitations}

\paragraph{Synthetic diagnostics}
All probing datasets are synthetic; their generation templates and few-shot examples are listed in Appendix~\ref{app:dataset_creation}, and domain experts confirmed their \emph{face validity} in Appendix~\ref{app:human_eval}.

\paragraph{Model scale and compute}
All experiments use three open models: Llama-3.1-8B, Qwen-2.5-7B, and Mistral-7B.  
For each ability we distributed the gradient-importance computation across eight NVIDIA A100 80 GB GPUs, which completed in about 25 minutes.  
The subsequent weight-zeroing step ran on a single A100 80 GB GPU and finished in roughly 5 minutes.  
Thus profiling one model over ten abilities plus downstream benchmark evaluation fits comfortably within a few GPU-hours.  
Profiling models beyond the 7--8B range may still require memory-efficient techniques such as gradient checkpointing.

\paragraph{Licensing and intended use}
The diagnostic datasets and code will be released under the MIT License for research and non-commercial use.  
They are not intended for high-stakes deployment or for ranking commercial systems without additional validation.

\paragraph{Documentation}
We provide full data statistics, generation templates, and class labels in Appendix~\ref{app:dataset_creation}.  A \texttt{README.md} with installation and reproduction scripts will accompany the code repository.

\section*{Ethics Statement}
\label{sec:ethics}

\paragraph{Data privacy and content}
All diagnostic items are generated from templated prompts and contain no personal or identifying details.  Volunteers manually screened a random sample and reported no offensive content (Appendix~\ref{app:human_eval}).

\paragraph{Benchmark licenses}
We rely only on benchmarks released under permissive licenses: ARC-Challenge, CommonsenseQA, GSM8K, HellaSwag, HumanEval, LogiQA, MBPP, Natural Questions Open, WinoGrande, and BIG-Bench Logical Deduction.  Our use remains within each dataset’s original research intent.

\paragraph{Synthetic artifact release}
To maintain anonymity during review, the diagnostic datasets, generation scripts, and validation labels will be placed in a public GitHub repository once the paper is accepted.  They will be distributed under the CC-BY-SA-4.0 license; accompanying code will use the MIT license.  The README file will specify intended research use and disclaim commercial deployment without additional validation.

\paragraph{Potential misuse}
Knowing how benchmarks decompose into abilities could, in theory, help actors craft adversarial tests or game leaderboard metrics.  We consider this risk low because reproducing our pipeline requires non-trivial compute, and transparency ultimately benefits the community by exposing hidden shortcuts.

\paragraph{Human subjects}
Ten adult volunteers participated in item validation.  No personal data were collected or stored beyond coarse demographics. Details are in Appendix~\ref{app:human_eval}.

\bibliography{anthology}

\appendix

\begin{table*}[t]
\centering
\small
\renewcommand{\arraystretch}{1.3} % Slightly increase row spacing
\resizebox{\linewidth}{!}{
\begin{tabular}{lcccccccccc}
\toprule[1.5pt] % Thicker top rule / double line effect
\multicolumn{11}{c}{\textbf{Llama-3.1-8B-Instruct (k=1.024\% MLP Ablation)}} \\
\midrule % Single line below the model info
\textbf{Ablated Ability} & \textbf{GSM8K} & \textbf{ARC-C} & \textbf{HellaSwag} & \textbf{LogiQA} & \textbf{BB Log.Ded.} & \textbf{WinoG.} & \textbf{CQA} & \textbf{NQ-Open} & \textbf{HumanEval} & \textbf{MBPP} \\
\midrule[1pt] % Slightly thicker line for main header separation / or \midrule followed by \midrule for double line
\textbf{Analogical Reasoning}   & 0.0833 & 0.0398 & 0.0337 & 0.0163 & 0.0114 & 0.0032 & 0.0426 & 0.0480 & 0.1006 & 0.1090 \\
\textbf{Commonsense Causal}     & 0.0583 & 0.0083 & 0.0337 & 0.0332 & 0.0439 & 0.0198 & 0.0256 & 0.0648 & 0.2730 & 0.1220 \\
\textbf{Contextual Recall}      & 0.0723 & 0.0229 & 0.0345 & 0.0210 & 0.0334 & 0.0168 & 0.0381 & 0.0884 & 0.1494 & 0.0922 \\
\textbf{Deductive Reasoning}    & 0.0750 & 0.0229 & 0.0273 & 0.0188 & 0.0678 & 0.0210 & 0.0486 & 0.0591 & 0.0469 & 0.0922 \\
\textbf{Inductive Reasoning}    & 0.0499 & 0.0338 & 0.0406 & 0.0254 & 0.0500 & 0.0337 & 0.0635 & 0.0193 & 0.1678 & 0.1316 \\
\textbf{Long-Term Knowledge}   & 0.0913 & 0.0083 & 0.0309 & -0.0098& 0.0466 & 0.0198 & 0.0364 & 0.0861 & 0.2657 & 0.1090 \\
\textbf{Quantitative Reasoning} & 0.0598 & 0.0116 & 0.0239 & -0.0031& 0.0692 & 0.0153 & 0.0472 & 0.0696 & 0.2272 & 0.1123 \\
\textbf{Semantic Relationship}  & 0.0872 & 0.0182 & 0.0330 & -0.0065& 0.0447 & -0.0016& 0.0256 & 0.0783 & 0.3275 & 0.1220 \\
\textbf{Spatial Reasoning}      & 0.0598 & 0.0291 & 0.0348 & 0.0059 & 0.0344 & 0.0092 & 0.0561 & 0.0942 & 0.1304 & 0.1156 \\
\textbf{Temporal Reasoning}     & 0.0441 & 0.0165 & 0.0309 & 0.0332 & 0.0457 & -0.0016& 0.0411 & 0.0936 & 0.2430 & 0.1220 \\
\bottomrule[1.5pt] % Thicker bottom rule / double line effect
\end{tabular}
}
\caption{Ability Impact Score (AIS) matrix for Llama-3.1-8B-Instruct across a curated suite of 10 benchmarks. Higher AIS values indicate greater performance loss upon ability ablation relative to the baseline's improvement over chance, suggesting higher dependence of the benchmark on that ability.}
\label{tab:ais_main_detailed}
\end{table*}

\section{Operationalized Ability Definitions and Diagnostic Task Principles}
\label{app:ability_definitions}

This appendix justifies the ten abilities used in \textsc{Benchmark Profiling}, situates each one within the Cattell--Horn--Carroll (CHC) model of intelligence \citep{carroll1993human,schneider2012cattell}, and explains how the corresponding synthetic diagnostic dataset was constructed.

Human cognition is \emph{distributed}: higher level skills co-recruit multiple lower level processes, and narrow processes are re-used across domains \citep{anderson2013architecture, oberauer2016control}.  CHC therefore models abilities as \emph{correlated but separable} factors rather than mutually exclusive boxes.  In the same spirit, our ten abilities were by design, chosen to be \emph{distinct enough} to yield interpretable weight profiles yet \emph{not so orthogonal} that they ignore real cognitive interactions.  Mild overlap is expected and even desirable: it lets our ablation analysis reveal which \emph{mixtures} of skills a benchmark rewards.  What matters empirically is that each diagnostic dataset is \emph{maximally diagnostic} for its target ability so that the gradient-importance procedure reliably tags a concentrated slice of parameters.  The robustness of the Ability-Impact profiles across three models (Section~\ref{sec:main_results}) supports this assumption.

Below, each ability entry follows the same template: (i)~cognitive-science grounding and CHC slot, (ii)~a motivating example, and (iii)~how the dataset was generated to isolate that skill.

\paragraph{Analogical Reasoning (CHC: \textit{Gf}-Induction).}
A proportional analogy such as \textit{bird:nest :: bee:?} demands mapping a relational schema rather than surface similarity; Raven's Progressive Matrices and related tasks tap the same faculty \citep{raven1938progressive,gentner1983structure,holyoak201213}. 

We authored four prompt templates that supply an \emph{A:B :: C:?} stem and four distractors.  Distractors are chosen by perturbing either A or B to share lexical or semantic features without preserving the relation (e.g., \textit{hive} (correct) vs.~\textit{honey, sting, wasp}).  This forces the model to attend to the latent mapping.

\paragraph{Commonsense \& Causal Reasoning (CHC: \textit{Gf} + script knowledge).}
Inferring that a neglected plant will wilt integrates causal schemas learned from everyday experience \citep{sloman2009causal, sap2020commonsense}.

Each question describes a three-to-five sentence vignette drawn from diverse domains (kitchen accidents, school routines, etc.).  We then ask for the most plausible cause or effect, sampling distractors from unrelated but thematically similar events to eliminate superficial cueing.  Scenarios were generated by large-model completion and manually filtered for obvious lexical shortcuts.

\paragraph{Contextual Recall (CHC: \textit{Gsm}).} Working-memory span underpins reading comprehension \citep{daneman1980individual,kane2002role}.

Two template families were used: (i)\emph{single-fact} passages of \mbox{2--3} sentences followed by a verbatim retrieval question, and (ii)\emph{multi-hop} passages of \mbox{4--6} sentences where the queried detail is the conjunction of two facts stated far apart.  All answers are extractive so no external knowledge is useful.

\paragraph{Deductive Reasoning (CHC: \textit{Gf}--Sequential Reasoning).}
Classical syllogisms illustrate rule-based deduction; accuracy correlates with measures of logical capacity \citep{johnson2001mental}.

Premises are generated by a symbolic template engine that instantiates first-order logic patterns (e.g., \emph{All S are P; No P are R; therefore ?}). Distractors violate exactly one rule to ensure that only a valid derivation succeeds.

\paragraph{Inductive Reasoning (CHC: \textit{Gf}--Induction).}
Discovering hidden regularities in sequences is central to hypothesis formation \citep{holland1986induction,lake2018generalization}.

We mine integer, geometric, and lexical pattern families (arithmetic progression, polygon naming, etc.).  For each instance we sample five in-context elements and ask for the sixth.  Distractors follow decoy rules (e.g., additive offset vs.~multiplicative) to penalize surface heuristics.

\paragraph{Long-Term Knowledge Recall (CHC: \textit{Glr}).}
Retrieving stored facts such as \emph{Canberra is Australia’s capital} maps to \textit{Glr} in CHC and has been probed extensively in LLMs \citep{petroni2019language,roberts2020much}.

We queried Wikidata for low-frequency entities, then generated four-choice trivia questions via a templating script.  We discard items whose answer string appears verbatim in the question to curb lexical leakage.

\begin{figure*}[h!]
\centering
\includegraphics[width=\linewidth]{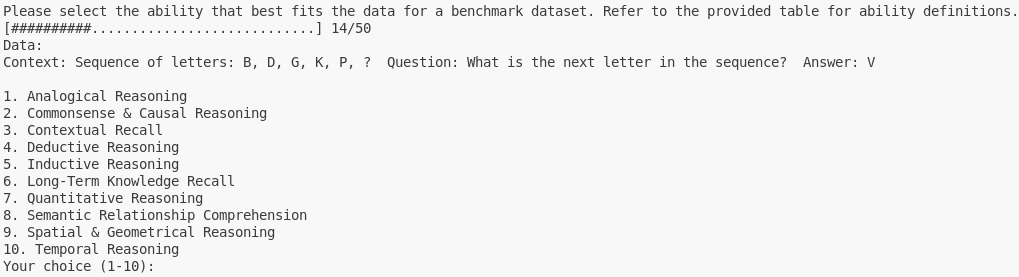}
\caption{Interface shown to volunteer experts during item validation.  Progress is indicated by a bar at the top.  Annotators read the prompt, inspect the ten ability options, and enter a numeric choice.}
\label{fig:expert_ui}
\end{figure*}

\paragraph{Quantitative Reasoning (CHC: \textit{Gq} + numeric \textit{Gf}).}
Multi-step word problems activate both quantitative knowledge and fluid reasoning \citep{cobbe2021training, lewkowycz2022solving}.

Templates embed 3--5 numbers, at least one of which is irrelevant, and require two operations (addition then division, etc.).  Distractor answers are produced by common student errors (off-by-one, wrong operator) as recommended by math-education literature \citep{sharma2019online}.

\paragraph{Semantic Relationship Comprehension (CHC: \textit{Gc}).}
Understanding taxonomical and role relations underlies lexical semantics \citep{miller1995wordnet, cummings2019clark}.

Each passage introduces 3--4 named entities in a mini-scenario (corporate hierarchy, biological taxonomy).  We ask about an implicit relationship (\emph{Who is Alice to Charlie?}) while distractors share topical words but break the relation type.

\paragraph{Spatial \& Geometrical Reasoning (CHC: \textit{Gv}).}
Textual spatial reasoning engages mental imagery and visuospatial sketchpad resources \citep{mani1982mental}.

We generate short descriptions of object layouts on a \(3\times 3\) grid and ask queries like \emph{Which object is directly below the circle?}.  Distractors include objects that are correct under mirror-flipped or rotated interpretations, so success requires consistent coordinate mapping.

\paragraph{Temporal Reasoning (CHC: sequencing facet of \textit{Gf}).}
Temporal sequencing develops early and is essential for narrative comprehension \citep{anderson2004integrated}.

Templates mention explicit times, durations, or adverbial order cues; questions ask which event came first, lasted longer, or overlapped.  Distractors are derived by permuting the true order.

\subsection*{Potential Overlaps and Taxonomy Limits}
Because CHC factors are \emph{correlated} (e.g., \textit{Gf} tasks share variance with \textit{Gsm}), perfect separation is neither feasible nor theoretically warranted \citep{tideman2004age,engle2004executive}.  We therefore aim for \emph{dominant association}: each diagnostic set should be \emph{most strongly} but not exclusively linked to its labeled ability.  Empirically, the distinct weight clusters uncovered by gradient analysis and the reproducible Benchmark Profiles across models (Figure~\ref{fig:js_similarity}) confirm that residual overlap does not hinder interpretability.  Indeed, partial overlap lets the Benchmark Profile capture real-world skill synergies, aligning with evidence that complex tasks draw on multiple CHC factors simultaneously \citep{schneider2012cattell}.

\section{Dataset Creation Prompts and Examples}
\label{app:dataset_creation}

The diagnostic datasets were generated using the \texttt{o4-mini-2025-04-16} API, guided by specific few-shot examples and instructions for each operationalized ability and task type. Table~\ref{tab:prompt_details} provides these details, used to generate 2000 examples per ability (split across task types for Contextual Recall). Output was requested in JSON format with keys: \texttt{context}, \texttt{question}, \texttt{options}, \texttt{answer}, \texttt{answer\_index}.

Table~\ref{tab:prompt_details} lists one representative template per ability along with the natural-language instructions fed to the API.  Each accepted completion was stored in a JSON schema with keys \texttt{context}, \texttt{question}, \texttt{options} (an array of four or five strings), \texttt{answer} (the correct option token), and \texttt{answer\_index} (zero-based integer).

\begin{table*}[!ht]
\centering
\small
\renewcommand{\arraystretch}{1.15}
\begin{tabularx}{\linewidth}{l X}
\toprule[1.5pt]

\multicolumn{2}{l}{\textbf{Base Model (no ablation)}} \\ \midrule[0.5pt]
\textbf{User} &
\emph{Charisma works for 8 hours every day. She has a timer to remind her to get up and walk for 5 minutes every hour she’s at work. After 5 days at the office, how many minutes has she walked?} \\[1.0ex]
\textbf{Assistant} &
To find out how many minutes Charisma walked, we need to calculate the total number of minutes she walked in 5 days.

Charisma walks 5 minutes every hour. Since she works 8 hours a day, she walks 5 * 8 = 40 minutes per day.

In 5 days, she will walk 40 * 5 = 200 minutes.

The answer is 200 minutes. \\ \midrule[1pt]

\multicolumn{2}{l}{\textbf{MLP-Only Ablation (fluent but weakened)}} \\ \midrule[0.5pt]
\textbf{User} & \emph{Same GSM8K question.} \\[1.0ex]
\textbf{Assistant} &
To find the total number of minutes Charisma walked, we need to calculate the total number of minutes she walked in 5 days.

Charisma works 8 hours a day and walks 5 minutes every hour.

First, we need to find the total number of minutes she works in a day.
8 hours * 60 minutes per hour = 480 minutes per day

Since she walks 5 minutes every hour, we multiply the total minutes she works in a day by 5.
480 minutes per day * 5 minutes per hour = 2400 minutes per day

Now, we need to find the total number of minutes she walked in 5 days.
2400 minutes per day * 5 days = 12000 minutes

Therefore, Charisma walked 12000 minutes in 5 days. \\ \midrule[1pt]

\multicolumn{2}{l}{\textbf{All-Layer Ablation (degraded)}} \\ \midrule[0.5pt]
\textbf{User} & \emph{Same GSM8K question.} \\[1.0ex]
\textbf{Assistant} &
Charisma works 8 hours a day, 5 days a week, so 5 * 8 = 40 hours.  40 * 60 = 2400 minutes.  2400 / 5 = 4800 minutes in 5 days.  8 * 60 = 4800 minutes in 5 days.  4800 / 5 = 9600 / 5 = 9600 / 5 = 9600 / 5 = 1920 5-minute walks, 5 minutes * 5 = 25 minutes per walk, 5 * 5 = 25.  25 * 5 = 1250, 1250 / 5 = 2500, 2500 / 5 = 500, 500 / 5 = 100, 100 / 5 = 20.  20 * 5 = 1000, 1000 / 5 = 200, 200 / 5 = 40, 40 / 5 = 8.  8 * 5 = 40, 40 / 5 = 8, 8 / 5 = 1.6, 1.6 * 5 = 8, 8 / 5 = 1.6, 1.6 * 5 = 8.  8 / 5 = 1.6, 1.6 * 5 = 8.  8 / 5 = 1.6, 1.6 * 5 = 8.  8 / 5 = 1.6,1.6 * 5 = 8.  8 / 5 = 1.6, 1.6 * 5 = 8.  8 / 5 = 1.6, 1.6 * 5 = 8.  8 / 5 = 1.6, 1.6 * 5 = 8.  8 / 5 = 1.6, 1.6 * 5 = 8.  8 / 5 = 1.6, 1.6 * 5 = 8.  8 / 5 = 1.6, 1.6 * 5 = 8.  8 / 5 = 1.6, 1.6 * 5 = 8. \\ \bottomrule[1.5pt]
\end{tabularx}
\caption{GSM8K test prompt evaluated by three model variants.  
The base model answers correctly, the MLP-only model remains fluent but over-counts,  
and the all-layer model degenerates into repetitive incoherence, illustrating  
why attention weights are left intact in our study.}
\label{tab:qualitative_single_turn}
\end{table*}

\section{Human Evaluation of Diagnostic Items}
\label{app:human_eval}

To verify that each synthetic question truly targets its intended skill, we invited ten independent volunteers to label a stratified sample of items drawn from the ten diagnostic datasets.

Table~\ref{tab:cognitive_abilities} was provided to the experts for reference.  Each volunteer saw fifty items (five from every ability) presented one at a time, as illustrated in Figure~\ref{fig:expert_ui}.  For every item they selected the single ability that best matched the question and flagged any unclear or sensitive content.

All annotators held at least a bachelor’s degree and were either postgraduate students or early-career researchers who responded to an internal mailing list. Participation was voluntary and unpaid. Three identified as women and seven as men, with ages ranging from 21 to 29. Annotators were drawn from institutions in Asia and North America.

\section{Detailed AIS Results and Raw Accuracies}
\label{app:detailed_benchmark_results}

Table~\ref{tab:ais_main_detailed} provides the detailed AIS matrix and raw accuracies for the baseline model \(\Theta\) and all 10 ability-ablated models \(\Theta^{a}\) at k=1.024\% across all evaluated benchmarks. These scores form the basis for creating the visualizations in Section~\ref{sec:main_results}.

\section{Qualitative Impact of Ablations}
\label{app:qualitative_mlp}

Table \ref{tab:qualitative_single_turn} contrasts the answers that the \textbf{Base}, \textbf{MLP-only}, and \textbf{All-layer} ablation versions of Llama-3.1-8B-Instruct give to the same GSM8K test question. The base model returns the correct total of \textbf{200 minutes}. The MLP-only model remains fluent but over-counts, replying with \textbf{12,000 minutes}. In the all-layer variant the response collapses into a repetitive numeric loop and never produces an answer. These qualitative differences align with the accuracy drops in Table \ref{tab:mlp_vs_all_layer_comparison} and underline why our study restricts ablation to MLP weights: they weaken targeted reasoning without destroying overall generation.

% --- Define Column Types ---
% C: Centered (Horizontally), Bold, Fixed Width p{} column (Vertically top-aligned)
\newcolumntype{C}[1]{>{\Centering\bfseries\arraybackslash}p{#1}}
% R: RaggedRight X column (Vertically top-aligned)
\newcolumntype{R}{>{\RaggedRight\arraybackslash}X}

% Reduce column separation slightly to help fit width
\setlength{\tabcolsep}{4pt}

\begin{table*}[!htbp]
\centering
\scriptsize % Use scriptsize for maximum space saving
\renewcommand{\arraystretch}{1.4} % Adjust row spacing slightly
% Use tabularx to fit \linewidth. Fixed p{} columns for first two, X for the rest.
% Adjusted widths: try smaller fixed widths for C columns. 4 Columns total now.
\begin{tabularx}{\linewidth}{|C{2.5cm}|C{2.5cm}|R|R|}

\hline
% Headers manually centered and bolded - 4 columns
\multicolumn{1}{|c|}{\textbf{Ability}} & \multicolumn{1}{c|}{\textbf{Task Type}} & \multicolumn{1}{c|}{\textbf{Representative Few-shot Example}} & \multicolumn{1}{c|}{\textbf{Instruction}} \\
\hline

% --- Analogical Reasoning ---
Analogical Reasoning & analogy & \texttt{Context}: 'Light : Dark :: Truth : ?' \newline \texttt{Question}: 'Which option best completes the analogy?' \newline \texttt{Options}: ['Lie', 'Fact', 'Shadow', 'Wisdom'] \newline \texttt{Answer}: 'Lie'. & Create a new analogy question. Use 'A : B :: C : ?' style or a similar analogical relationship in context, and ask which option completes it. Output as JSON with the required fields. \\
\hline

% --- Commonsense & Causal ---
Commonsense \& Causal Reasoning & everyday cause effect & \texttt{Context}: 'Tom left his ice cream in the sun on a hot day.' \newline \texttt{Question}: 'What likely happened to the ice cream?' \newline \texttt{Options}: ['It melted', 'It froze', 'It caught fire', 'It grew larger'] \newline \texttt{Answer}: 'It melted'. & Now write a new commonsense cause-and-effect question. The context should be a simple scenario, and the question asks for a logical outcome or reason. Ensure the answer is based on everyday common sense. Output in JSON format. \\
\hline

% --- Contextual Recall ---
% Use multirow only for the Ability name, spanning 2 rows.
\multirow{2}{*}{\Centering\bfseries Contextual Recall} & single fact recall & \texttt{Passage}: 'In an ancient manuscript... Nile flows through Egypt while the Tiber traverses Rome.' \newline \texttt{Question}: 'Based on the passage, which river is associated with Rome?' \newline \texttt{Options}: ['Nile', 'Tiber', 'Danube', 'Ganges'] \newline \texttt{Answer}: 'Tiber'. & Using the context... generate one new, diverse and non-redundant MCQ. Output valid JSON object with keys: \texttt{context}, \texttt{question}, \texttt{options}, \texttt{answer}, \texttt{answer\_index}. Use varied domains. \\
\cline{2-4} % Partial line between the two task types, spans cols 2-4 now
 & multi hop recall & \texttt{Passage}: 'At a Geneva symposium... 1969 lunar landing spurred robotics. Later in Tokyo... noted automation...' \newline \texttt{Question}: 'Which event... linked to inspiration for robotics?' \newline \texttt{Options}: ['Geneva...', '1969 lunar landing', 'Tokyo...', 'Automation...'] \newline \texttt{Answer}: '1969 lunar landing'. & Now generate a new multi-hop question. Passage should blend facts, question require combination. Present JSON ensuring multi-hop reasoning. \\
\hline

% --- Deductive Reasoning ---
Deductive Reasoning & logic puzzle & \texttt{Passage}: 'If every cat... black animals are calm... Whiskers is in the room.' \newline \texttt{Question}: 'Based on the passage, what can we deduce about Whiskers?' \newline \texttt{Options}: ['Whiskers is calm.', 'Whiskers is not black.', ...] \newline \texttt{Answer}: 'Whiskers is calm.'. & Now create a new deductive reasoning question. Provide context with premises/clues, question requires deducing answer. Output JSON. \\
\hline

% --- Inductive Reasoning ---
Inductive Reasoning & pattern completion & \texttt{Sequence}: 'A, C, E, G, ?' \newline \texttt{Question}: 'What is the next letter...?' \newline \texttt{Options}: ['H', 'I', 'J', 'K'] \newline \texttt{Answer}: 'I'. & Generate new inductive reasoning question based on pattern. Ensure question asks for next element/rule. Provide answer/\texttt{answer\_index} JSON. \\
\hline

% --- Long-Term Knowledge ---
Long-Term Knowledge Recall & world fact & \texttt{Context}: 'This question is about world geography.' \newline \texttt{Question}: 'What is the capital city of Australia?' \newline \texttt{Options}: ['Sydney', 'Canberra', 'Melbourne', 'Perth'] \newline \texttt{Answer}: 'Canberra'. & Create new world-knowledge question. Provide brief context if needed, question must be answered from general knowledge. Ensure JSON format. \\
\hline

% --- Quantitative Reasoning ---
Quantitative Reasoning & arithmetic word problem & \texttt{Context}: 'Alice had 5 apples. She gave 2 to Bob and then bought 3 more.' \newline \texttt{Question}: 'How many apples does Alice have now?' \newline \texttt{Options}: ['6', '5', '8', '10'] \newline \texttt{Answer}: '6'. & Generate new math word problem/quantitative question. Context provides numbers/scenario, question asks for result. Provide answer/\texttt{answer\_index} JSON. \\
\hline

% --- Semantic Relationship ---
Semantic Relationship & roles and relations & \texttt{Passage}: 'Alice is Bob's mother. Bob is Charlie's teacher.' \newline \texttt{Question}: 'Who is Alice to Charlie?' \newline \texttt{Options}: ['His mother', 'His teacher', 'His grandmother', 'Not related'] \newline \texttt{Answer}: 'His grandmother'. & Generate new passage and question about relationships/roles. Passage contains >= 2 entities with relationship. Ask question testing understanding. Output JSON. \\
\hline

% --- Spatial Reasoning ---
Spatial Reasoning & spatial relation & \texttt{Context}: 'There is a triangle to the left of a square, and a circle above the triangle.' \newline \texttt{Question}: 'Which shape is directly below the circle?' \newline \texttt{Options}: ['Triangle', 'Square', 'Circle', 'None'] \newline \texttt{Answer}: 'Triangle'. & Generate new spatial reasoning question. Context: description of locations or simple geometry. Ask about relative position, direction, or basic inference. Provide JSON output. \\
\hline

% --- Temporal Reasoning ---
Temporal Reasoning & temporal order & \texttt{Context}: 'John's meeting started at 9:00 AM... lasted 2 hours. Mary's meeting started at 10:30 AM...' \newline \texttt{Question}: 'Whose meeting ended later?' \newline \texttt{Options}: ['John', 'Mary', 'Same time', 'Not enough info'] \newline \texttt{Answer}: 'John'. & Now create new temporal reasoning question. Context with >= 2 events/time points. Ask about order/timing (e.g., first, duration). Output JSON. \\
\hline

\end{tabularx}
\caption{Few-shot Examples and Instructions for Diagnostic Dataset Generation.}
\label{tab:prompt_details}
\end{table*}

\end{document}